\documentclass[letterpaper, 10 pt, conference]{ieeeconf}  

\IEEEoverridecommandlockouts                              

\usepackage{multirow}
\usepackage{lipsum}
\usepackage{amsmath}
\usepackage{amssymb}
\usepackage[ruled,vlined]{algorithm2e}
\usepackage{graphicx}
\graphicspath{{./Figures/}}
\usepackage{cite}
\usepackage{amsmath,amssymb,amsfonts}
\usepackage{algorithmic}
\usepackage{graphicx}
\usepackage{textcomp}
\usepackage[dvipsnames]{xcolor}
\usepackage{subfig}
\usepackage[font=small,skip=0pt]{caption}
\usepackage{dblfloatfix}
\usepackage{geometry}
\usepackage{multirow}
\usepackage{xcolor}
\usepackage{comment}
\usepackage{booktabs}
\usepackage[normalem]{ulem}
\def\BibTeX{{\rm B\kern-.05em{\sc i\kern-.025em b}\kern-.08em
    T\kern-.1667em\lower.7ex\hbox{E}\kern-.125emX}}

\usepackage{scalerel}
\usepackage{tikz}
\usetikzlibrary{svg.path}

\definecolor{orcidlogocol}{HTML}{A6CE39}
\tikzset{
  orcidlogo/.pic={
    \fill[orcidlogocol] svg{M256,128c0,70.7-57.3,128-128,128C57.3,256,0,198.7,0,128C0,57.3,57.3,0,128,0C198.7,0,256,57.3,256,128z};
    \fill[white] svg{M86.3,186.2H70.9V79.1h15.4v48.4V186.2z}
                 svg{M108.9,79.1h41.6c39.6,0,57,28.3,57,53.6c0,27.5-21.5,53.6-56.8,53.6h-41.8V79.1z M124.3,172.4h24.5c34.9,0,42.9-26.5,42.9-39.7c0-21.5-13.7-39.7-43.7-39.7h-23.7V172.4z}
                 svg{M88.7,56.8c0,5.5-4.5,10.1-10.1,10.1c-5.6,0-10.1-4.6-10.1-10.1c0-5.6,4.5-10.1,10.1-10.1C84.2,46.7,88.7,51.3,88.7,56.8z};
  }
}

\newcommand\orcidicon[1]{\href{https://orcid.org/#1}{\mbox{\scalerel*{
\begin{tikzpicture}[yscale=-1,transform shape]
\pic{orcidlogo};
\end{tikzpicture}
}{|}}}}

\usepackage{hyperref} 

\title{\LARGE \bf Optimising CSRNet with parameter-free attention mechanisms for crowd counting in public transport}

\author{Aida Rostamza\orcidicon{0009-0002-2036-2191} \textit{Graduate Student Member, IEEE}, Enrico Del Re\orcidicon{0009-0002-6417-5902} \textit{Graduate Student Member, IEEE}, \\ Joshua Cherian Varughese\orcidicon{0000-0002-3250-0742} \textit{Member, IEEE} and Cristina Olaverri-Monreal\orcidicon{0000-0002-5211-3598} \textit{Senior Member, IEEE}%
\thanks{ Johannes Kepler University Linz, Austria, Department Intelligent Transport Systems, Altenberger Straße 69, 4040 Linz, Austria.
\texttt{\{aida.rostamza, enrico.del\_re, joshua.varughese, cristina.olaverri-monreal\}@jku.at}}%
}

\begin{document}
\newcommand\mc[1]{\multicolumn{1}{c}{#1}} 
	
	\maketitle
	\thispagestyle{empty}
	\pagestyle{empty}
	
	\begin{abstract}
Occupancy estimation and crowd counting are critical tasks in designing smart and efficient public transport vehicles. Given that public transport loading can vary from sparse to crowded, classical models for occupancy estimation must be adapted to suit this purpose. Attention mechanisms have shown remarkable capability in enhancing the representational power of deep neural networks for crowd counting in congested scenes with occlusion, complex backgrounds, and perspective distortion. However, conventional approaches, often implemented as parameterized sub-networks within convolutional layers, inevitably increase model size and computational cost, limiting deployment on resource-constrained edge devices. This paper investigates the effectiveness of state-of-the-art parameter-free attention mechanisms for crowd counting and density map estimation in highly congested scenes. We evaluate channel-wise (PFCA), spatial-wise (SA), and 3-D (SimAM) modules and compare their performance with parameterized attention modules constrained to introduce no more than 1\% additional parameters. Furthermore, we present a novel combination of attention mechanisms that combines the strengths of PFCA and SA (PFCASA) customized for analyzing video streams onboard public transport systems. Using CSRNet as the backbone, experiments on the ShanghaiTech dataset demonstrate that parameter-free attention mechanisms achieve comparable or superior accuracy without introducing additional model parameters. A detailed performance analysis further reveals that PFCASA outperforms other attention modules in scenes with fewer than 40 individuals, while PFCA shows greater effectiveness as crowd density increases, underscoring their potential applicability for integration into smart public transport modalities.
	\end{abstract}
	
\section{Introduction}\label{sec:intro}

In recent years, crowd analysis in densely populated environments has gained increasing attention for public safety, motivated by tragic incidents such as the 2015 Shanghai stampede~\cite{Shanghai_2015} and the 2022 Seoul crowd crush~\cite{SON2025106741}. Beyond safety concerns, crowd counting also plays a crucial role in traffic management and urban planning, such as directing passengers to less crowded metro platforms, which can increase transport capacity by up to 54.8\% during peak hours~\cite{Cristina_optimization,zhou2015optimization}. However, achieving reliable crowd counting in dense scenes remains challenging due to severe occlusion, perspective distortion, and scale variation, as well as the high computational cost of real-time implementation on edge devices. 

Since the success of CSRNet~\cite{Li_csrnet}, many crowd-counting methods have adopted VGG-16 or VGG-19 as their encoder backbone, while focusing primarily on designing new decoder architectures to improve performance. As a result, research has largely emphasized architectural modifications rather than investigating the effectiveness of feature enhancement components within a fixed baseline~\cite{li2024combinatorial}. In particular, attention mechanisms are often introduced as auxiliary modules to improve performance, but their individual contributions and efficiency are rarely systematically evaluated under controlled conditions. Furthermore, most existing attention modules rely on additional learnable parameters, increasing model complexity. Parameter-free attention mechanisms, which enhance feature representation without introducing additional parameters, remain largely overlooked in crowd counting. Consequently, it remains unclear whether performance improvements stem from the attention mechanism itself or simply from increased model capacity, and whether parameter-free alternatives can provide comparable benefits more efficiently.

To address this gap, this study investigates the integration of parameter-free attention mechanisms into the CSRNet baseline~\cite{Li_csrnet}. The goal is to evaluate the effectiveness of these modules in enhancing crowd counting performance and to explore their potential integration into practical public transport occupancy detection models. We hypothesize that parameter-free attention can improve feature representation and counting accuracy without increasing model complexity, enabling efficient deployment on edge devices in public transport settings. Specifically, we evaluate the effectiveness of parameter-free channel attention (PFCA)~\cite{shi2023pfca}, spatial attention (SA)~\cite{Wang_sa}, their combination in a novel PFCASA attention mechanism, 3-D attention (SimAM)~\cite{yang2021simam}, and compare them with parameterized attention modules constrained to introduce no more than 1\% additional parameters, as well as the original CSRNet implementation, on the ShanghaiTech dataset~\cite{Shanghaitech}. This constraint is motivated by the fact that CSRNet contains approximately 16.26 million parameters; thus, a 1\% increase corresponds to roughly 163,000 additional parameters. While this increase appears marginal in large-scale networks, it represents a substantial addition for lightweight models operating in the range of thousands to hundreds of thousands of parameters. Our experiments focus on Part B of the dataset, which is more representative of public transport environments due to its moderate density and varied crowd distribution.

The remainder of this paper is organized as follows: \autoref{sec:related} reviews related work on crowd counting and attention mechanisms. \autoref{sec:background} explains the CSRNet architecture and the attention modules of interest. \autoref{sec:method} describes the dataset, training details and experimental setup. \autoref{sec:results} and \autoref{sec:discussion} present the results and the discussion, respectively. Finally, \autoref{sec:conclusion} concludes the paper.
	

\section{Related Work} \label{sec:related}

Crowd counting methods are generally divided into detection-based, regression-based, and density-based approaches. Detection-based methods employ object detectors to locate and count individuals but struggle in dense crowds due to occlusion and overlap. Regression-based methods instead learn a direct mapping from image features to counts, avoiding detection; however, their reliance on handcrafted features and weak spatial context modeling limits their robustness in complex scenes~\cite{li2021approaches}. To address these limitations, density-based methods formulate crowd counting as a density map estimation problem~\cite{Lempitskydamp}. In this approach, the total count is obtained by integrating over a predicted density map, allowing the model to capture both the number and spatial distribution of individuals. With the advancement of convolutional neural networks (CNNs), density-based methods have become the dominant paradigm, achieving significant improvements in accuracy and robustness in congested scenes~\cite{Rodriguez_dmap}.

Following the success of CSRNet, most crowd-counting approaches adopt VGG-16 or VGG-19 as their backbone. Typically, VGG is used as the encoder, while research focuses on improving the decoder design, as seen in Soft-CSRNet~\cite{Bakour_SoftCSRNet}, CSRNet+~\cite{Zhao_CSRnet}, and CDNet~\cite{Zhao_cdnet}. Similarly, SANet~\cite{cao2018scale} and TEDnet~\cite{jiang2019crowd} propose alternative encoder–decoder architectures to enhance feature aggregation and improve density estimation accuracy. In addition to encoder–decoder architectural design, a wide range of approaches have been proposed to address challenges such as scale variation, occlusion, and background noise, including multi-column architectures such as MCNN~\cite{Zhang_mcnn} and patch-based methods like Switching-CNN~\cite{Sam_switching}. More recently, attention-based models, including MSAN~\cite{Yang_msan}, SCAR~\cite{Gao_scar}, and SFANet~\cite{Zhu_sfanet}, have further improved performance by enhancing feature representation, enabling models to focus on informative regions while reducing the influence of irrelevant background.

Despite these advances, several limitations remain. Multi-column architectures substantially increase the number of parameters and computational cost, making end-to-end training more difficult. Patch-based methods often suffer from boundary artifacts and inconsistent local predictions, while attention-based models introduce additional learnable parameters, increasing model complexity, memory usage, and power consumption, which limits their suitability for edge deployment.~\cite{li2021approaches}


\section{Model Overview} \label{sec:background}

In this section, we provide an overview of the model and components adopted in our study. Specifically, we describe the CSRNet architecture, which serves as the backbone of our framework, as well as the attention modules integrated to investigate their impact on model performance.

\begin{figure*} [!t]%
\centering
\includegraphics[width=0.75\textwidth]{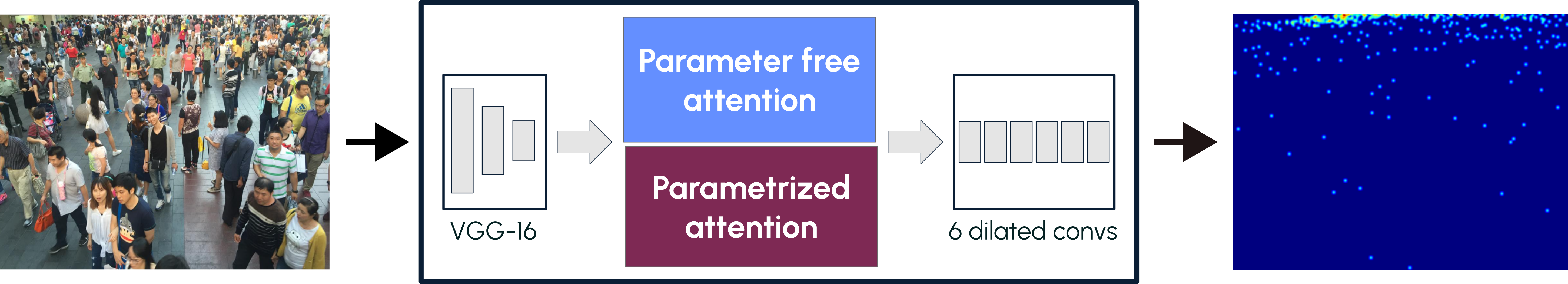}
\vspace{2mm}
\caption{Figure shows the setup used for investigating the impact of integrating attention modules between the frontend (VGG-16) and the dilated convolutions in the backend of the CSRNet. A total of eleven attention configurations as shown in \autoref{tab:results} were tested, comparing their performance with the original CSRNet.}
\label{fig:eval_configs}
\end{figure*}%

\subsection{CSRNet}
CSRNet~\cite{Li_csrnet} is a single-column CNN architecture designed for crowd counting and density estimation in highly congested scenes. It utilizes the first ten convolutional layers of VGG-16 as a frontend for feature extraction and six dilated convolutional layers as a backend for generating high-quality density maps. 
By demonstrating the redundancy of learned features across MCNN's columns~\cite{Li_csrnet}, CSRNet initiated the shift toward efficient single-column designs. Leveraging dilated convolutions to enlarge the receptive field without reducing spatial resolution and employing a straightforward training procedure, it established a foundational baseline for subsequent research. 
 
\subsection{Attention mechanisms}
\subsubsection{Parameter-Free Channel Attention Module (PFCA)}

The PFCA is a channel attention module that enhances feature representation by reweighting channel importance based on statistical information without introducing additional parameters.

Given an input feature map $X \in \mathbb{R}^{N \times C \times H \times W}$, where $N$, $C$, and $H \times W$ denote the batch size, number of channels, and spatial dimensions respectively, PFCA first computes the channel vector $U \in \mathbb{R}^{N \times C \times 1 \times 1}$ by applying global average pooling across the spatial dimensions. For each channel $j$, the attention weight is obtained as:

\begin{equation}
V_j = \frac{(U_j - \mu)^2 + 2(\sigma^2 + \lambda)}{4(\sigma^2 + \lambda)}
\end{equation}
where $\mu$ and $\sigma^2$ represent the mean and variance of the channel vector $U$ computed in the channel dimension, and $\lambda$ is a small constant (e.g., $10^{-4}$) for numerical stability. The refined feature map is then computed as:

\begin{equation}
Y = X \odot \text{Sigmoid}(V)
\end{equation}

where $\odot$ denotes element-wise multiplication, and $\text{Sigmoid}(\cdot)$ is the sigmoid activation function.

\subsubsection{Parameter-Free Spatial Attention Module (SA)}

The SA module models spatial relations, allowing the network to focus on salient regions within feature maps without increasing the number of parameters. Following~\cite{Wang_sa}, given an input feature map $X \in \mathbb{R}^{C \times H \times W}$, where $C$ is the number of channels and $H \times W$ are the spatial dimensions, the spatial attention weight at position $(i, j)$ is computed as:

\begin{equation}
p(i,j) = \text{Softmax}\left(\sum_{k=1}^{C} x_k(i,j)\right)
\end{equation}
The reweighted feature map is obtained by:
\begin{equation}
X_k(i,j) = x_k(i,j) \odot p(i,j)
\end{equation}

where $\odot$ denotes element-wise multiplication, and $\text{Softmax}(\cdot)$ is the softmax activation function applied over the spatial domain to produce normalized attention weights.
\subsubsection{Parameter-Free 3-D Attention Module (SimAM)}

SimAM introduces a parameter-free 3-D attention mechanism that jointly models spatial and channel dependencies. Given the input feature map $X \in \mathbb{R}^{C \times H \times W}$, where $C$ is the number of channels and $H \times W$ are the spatial dimensions, SimAM models the linear separability between a target neuron and other neurons within the same channel as an energy function, and derives a closed-form solution for the minimal energy as follows:

\begin{equation}
e_t^* = \frac{4(\hat{\sigma}^2 + \lambda)}{(t - \hat{\mu})^2 + 2\hat{\sigma}^2 + 2\lambda}
\end{equation}

where $\hat{\mu} = \frac{1}{M}\sum_{i=1}^{M} x_i$ and $\hat{\sigma}^2 = \frac{1}{M}\sum_{i=1}^{M}(x_i - \hat{\mu})^2$ are the mean and variance of all activations within a channel ($M = H \times W$), $t$ is the target neuron's activation value, and $\lambda$ is a regularization parameter (typically set to $10^{-4}$). 
Neurons with lower $e_t^*$ are considered more distinctive and important; thus, attention is inversely proportional to $e_t^*$. 
The refined feature map is obtained as:

\begin{equation}
\tilde{X} = X \odot \text{Sigmoid}\left(\frac{1}{E}\right)
\end{equation}

where $ \odot $ denotes element-wise multiplication and $\text{Sigmoid}(\cdot)$ represents the sigmoid activation function, with $E$ being the set of all $e_t^*$ across spatial and channel dimensions.

\subsubsection{Parameter-Free Channel and Spatial Attention (PFCASA)}
Inspired by the Convolutional Block Attention Module (CBAM)~\cite{woo2018cbam}, which sequentially applies channel and spatial attention to refine feature representations, we propose a Parameter-Free Channel and Spatial Attention (PFCASA) module. Unlike CBAM, which relies on learnable parameters in both its channel and spatial attention components, the proposed PFCASA replaces them with PFCA and SA, respectively. This design improves feature representation without introducing additional learnable parameters.

Given an intermediate feature map $ X \in \mathbb{R}^{C \times H \times W} $, where $ C $ is the number of channels and $ H \times W $ denotes the spatial dimensions, the channel attention output $ X_c $ and the channel–spatial attention output $ X_{cs} $ of PFCASA can be expressed as:

\begin{equation}
X_c = PFCA(X) \odot X
\end{equation}
\begin{equation}
X_{cs} = SA(X_{c}) \odot (X_{c})
\end{equation}

where $PFCA(\cdot)$ and $ SA(\cdot)$ denote the parameter-free channel and spatial attention modules, respectively, and $ \odot $ represents element-wise multiplication.

\subsubsection{Parameterzied Attention Modules}
To investigate the effectiveness of parameter-free attention mechanisms as efficient alternatives, we compare them with several widely used parameterized attention modules. Specifically, Squeeze-and-Excitation (SE)\cite{hu2018squeeze} is selected as a representative channel attention mechanism, while Coordinate Attention (CAM) \cite{Hou_cam} and Convolutional Block Attention Module (CBAM) are included as hybrid attention mechanisms incorporating both spatial and channel information.

As described in \autoref{sec:intro}, the parameterized attention modules are evaluated under a strict 1\% parameter constraint relative to the CSRNet baseline. This design allows the contribution of the attention mechanisms to be assessed with minimal influence from increased parameter capacity. To analyze the impact of parameterization levels, two configurations of each parameterized attention module are evaluated using different reduction ratios. The first configuration uses a reduction ratio of 16, resulting in a minimal parameter increase relative to the baseline. The second configuration uses the minimum reduction ratio that produces a parameter increase closest to the defined constraint. This design enables a controlled comparison between parameterized and parameter-free attention mechanisms under equivalent parameter constraints.

\section{Methodology} \label{sec:method}

In this section, we present the methodology used to evaluate the integration of parameter-free attention mechanisms into the CSRNet baseline and to assess our hypothesis. \autoref{fig:eval_configs} illustrates the overall configurations evaluated in this study.

\subsection{Dataset Description}

\begin{figure}
    \centering
    \includegraphics[width=0.75\linewidth]{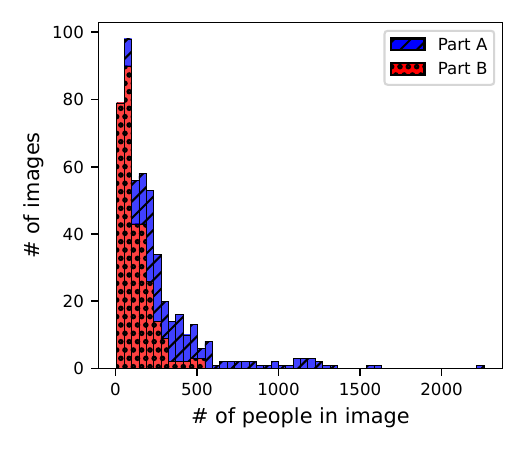}
    \caption{Histogram showing the distribution of the number of people in images in the test split of ShanghaiTech Part A and B. Part A is stacked on top of part B}
    \label{fig:dataset_hist}
\end{figure}

The ShanghaiTech dataset \cite{Shanghaitech} is a widely used benchmark for crowd counting and density estimation. It contains 1,198 annotated images with a total of 330,165 individuals, captured from diverse scenes across Shanghai. The dataset is divided into two parts: Part A, which consists of 482 images collected from the Internet featuring highly congested urban scenes with dense crowds, and Part B, which includes 716 images captured on the streets of Shanghai with relatively sparse crowds. The histogram in \autoref{fig:dataset_hist} shows the distribution of the number of images in the dataset with varying number of people per image.

\subsection{Training details}
The training setup follows the original CSRNet configuration in~\cite{Li_csrnet}. The first ten convolutional layers are fine-tuned from a pre-trained VGG-16 model, while the remaining layers are initialized using a Gaussian distribution with a standard deviation of 0.01. Training is performed using stochastic gradient descent (SGD) with a weight decay of $5 \times 10^{-4}$, momentum of 0.95, and a fixed learning rate of $1 \times 10^{-6}$.

Ground-truth density maps are generated using geometry-adaptive kernels defined as:
\begin{equation}
F(x) = \sum_{i=1}^{N} \delta(x - x_i) * G_{\sigma_i}(x), \quad \text{where } \sigma_i = \beta \bar{d}_i
\end{equation}

For each annotated target object $x_i$ in the ground truth $\delta$, $d_i$ is computed as the average distance to its $k$ nearest neighbors. The density map is then generated by convolving $\delta(x - x_i)$ with a Gaussian kernel with parameter $\sigma_i$ (standard deviation). Following ~\cite{Li_csrnet} we employ a geometry-adaptive kernel with $\beta = 0.3$ and $k = 3$ for Part~A, while a fixed Gaussian kernel with $\sigma = 15$ is used for Part B.


The Euclidean distance is employed as the loss function to measure the difference between the predicted and ground-truth density maps:
\begin{equation}
L(\Theta) = \frac{1}{2N} \sum_{i=1}^{N} | Z(X_i; \Theta) - Z_i^{GT} |_2^2
\end{equation}
where $N$ is the batch size, $Z(X_i; \Theta)$ is the predicted density map for input image $X_i$, and $Z_i^{GT}$ is the corresponding ground-truth density map.

Following the original paper, we applied the same data augmentation strategy. From each image, nine patches of one-quarter the original size were cropped. The first four patches correspond to non-overlapping quadrants, while the remaining five are randomly sampled from the image. Each patch is then horizontally flipped, effectively doubling the size of the training set.

\subsection{Evaluation}\label{sec:experiment}

To assess the impact of integrating parameter-free attention modules, we tested several configurations as shown in (\autoref{fig:eval_configs}). We hypothesized that placing attention modules between the frontend and backend allows the backend to capture more relevant spatial and contextual information, thereby improving counting accuracy. To validate this, we conducted an ablation study evaluating PFCA, SA, SimAM and PFCASA and compared them to some widely used parameterized attention modules including CAM, SE, CBAM and the original CSRNet baseline. 

All experiments were performed on the ShanghaiTech dataset \cite{Shanghaitech} for 400 epochs, and the best results out of three runs under identical settings were reported.

To assess the model’s performance, the Mean Absolute Error (MAE), Mean Squared Error (MSE) and Accuracy evaluation metrics were adopted and defined as follows:

\begin{equation}
\text{MAE} = \frac{1}{N} \sum_{i=1}^{N} \left| C_i - C_i^{\text{gt}} \right|   
\label{eq_mae}
\end{equation}

\begin{equation}
\text{MSE} = \sqrt{\frac{1}{N} \sum_{i=1}^{N} (C_i - C_i^{\text{gt}})^2} 
\label{eq_mse}
\end{equation}

\begin{equation}
\text{Accuracy} = 1 - \frac{1}{N} \sum_{i=1}^{N} \frac{\left| C_i - C_i^{\text{gt}} \right|}{C_i^{\text{gt}}}
\label{eq_acc}
\end{equation}

where $N$ is the total number of test images, $C_i$ is the predicted crowd count for image $i$, and $C_i^{gt}$ is the corresponding ground truth count. MAE primarily measures the accuracy of the predictions, while accuracy is meant to normalize the error computed by MAE according to the ground truth.

It should be noted that although the training procedure was consistent across all configurations, the SA module failed to train. Upon investigation, this failure was found to be due to the use of the softmax operation in the original SA implementation. 

To address this issue, we replaced softmax with a sigmoid activation, which restored the model’s learning capability and yielded performance comparable to the baseline. Consequently, all subsequent experiments comparing SA with other attention modules were conducted using the sigmoid-based variant rather than the softmax baseline.

\begin{table}[!t]
\centering
\caption{Results of integrating attention mechanisms into the CSRNet baseline on ShanghaiTech test sets.}
\label{tab:results}
\setlength{\tabcolsep}{3pt}
\renewcommand{\arraystretch}{1.1}
\begin{tabular}{|l|c|c|c|c|c|c|}
\hline
\multirow{2}{*}{\textbf{Attention}} & \multicolumn{2}{c|}{\textbf{Part A}} & \multicolumn{2}{c|}{\textbf{Part B}} & \multirow{2}{*}{\textbf{Params(K)}} & \multirow{2}{*}{\textbf{Added Params}} \\
\cline{2-5}
& MAE & MSE & MAE & MSE & & \\
\hline
No Attn.     & 68.20  & 115.00 & 10.60 & 16.00 & 16,263,041  & No  \\
PFCA         & 73.36  & 113.98 & 8.50  & 12.78 & 16,263,041  & No  \\
SA           & 80.44  & 117.43 & 9.07  & 13.97 & 16,263,041  & No  \\
PFCASA       & 73.25  & 108.34 & 9.25  & 14.15 & 16,263,041  & No  \\
SimAM        & 77.08  & 116.47 & 9.24  & 14.40 & 16,263,041  & No  \\
SE(r=4)      & 79.06  & 117.73 & 8.87  & 14.55 & 16,394,561  & Yes \\
CAM(r=8)     & 83.80  & 121.33 & 8.46  & 15.02 & 16,363,009  & Yes \\
CBAM(r=4)    & 71.85  & 114.06 & 8.85  & 13.63 & 16,394,659  & Yes \\
SE(r=16)     & 81.35  & 119.36 & 8.67  & 14.67 & 16,296,257  & Yes \\
CAM(r=16)    & 83.80  & 124.92 & 8.42  & 13.45 & 16,313,761  & Yes \\
CBAM(r=16)   & 74.09  & 118.30 & 9.53  & 14.39 & 16,296,355  & Yes \\
\hline
\end{tabular}
\end{table}

\section{Results}\label{sec:results}

\subsection{Importance of attention mechanisms}

Table~\ref{tab:results} summarizes the impact of integrating different attention mechanisms into the CSRNet baseline on the ShanghaiTech dataset. Since this work focuses on moderate-density crowd scenarios relevant to public transport applications, the discussion primarily emphasizes the results on Part B.

On ShanghaiTech Part B, the baseline CSRNet achieves an MAE of 10.60. All evaluated attention mechanisms improve performance over the baseline, confirming the importance of feature enhancement even within a strong fixed architecture. Among the parameter-free attention methods, PFCA achieved the best overall performance with an MAE of 8.50, corresponding to a relative error reduction of approximately 19.81\% compared to the baseline. SA and SimAM also improved accuracy, achieving MAEs of 9.07 and 9.24, respectively. These results demonstrate that parameter-free attention mechanisms can significantly enhance feature representation without introducing additional parameters.

Parameterized attention mechanisms also provided improvements, but their performance remains comparable to parameter-free alternatives. Specifically, CBAM($r$=4) and SE($r$=4) achieve MAEs of 8.85 and 8.87, respectively, while SE($r$=16) achieves an MAE of 8.67. Although these modules improve accuracy, they introduce additional parameters ranging from 0.20\% to 0.81\%. In contrast, PFCA achieves the best performance overall while adding no parameters, highlighting its superior efficiency–accuracy trade-off.

\subsection{Accuracy at different densities}

\begin{figure}[!t]
\centering
\subfloat[][]{\includegraphics[width=0.90\linewidth]{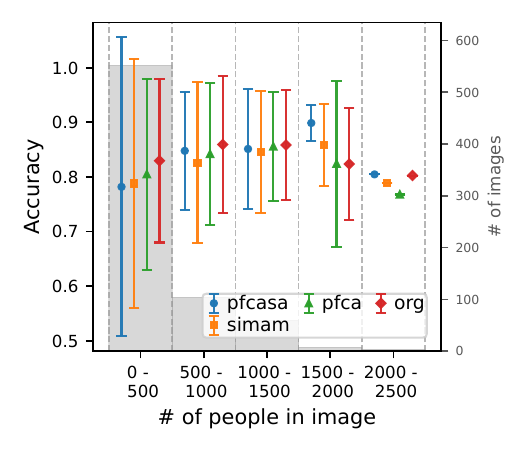}
\label{fig:partA_full}}
\hfil
\subfloat[][]{\includegraphics[width=0.90\linewidth]{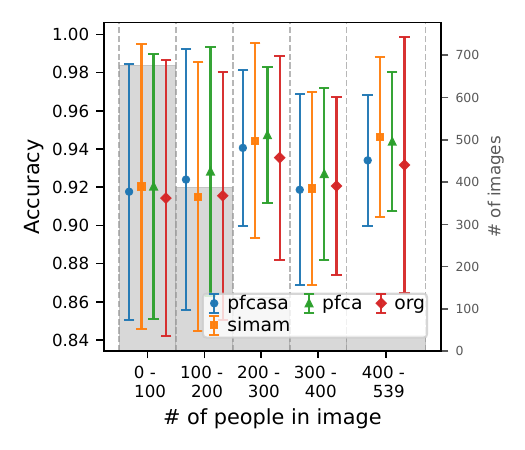}
\label{fig:partB_full}}
\caption{Figures \protect\subref{fig:partA_full} and \protect\subref{fig:partB_full} show the accuracy comparison at different crowd densities between different configurations tested for Part A and Part B of the Shanghai Tech dataset respectively.
Additionally, the background histogram indicates the number of test images in each density bin, giving context for how the results aggregate into the overall accuracy and MAE.}
\label{fig:partAB}
\end{figure}

Figure \ref{fig:partAB} illustrates the accuracy of the models on binned sections of the datasets. For Part A, shown in Figure \autoref{fig:partA_full}, the original model  outperforms the different attention mechanisms across most bins. For Part B in Figure \autoref{fig:partB_full} the different attention mechanisms show higher accuracy across the majority of the dataset, with different mechanisms performing better or worse depending on the density of people.

For 0 - 20 people per image, the original network achieves an accuracy of 0.882$\pm$0.088 while PFCASA achieves an accuracy of 0.887$\pm$0.088. For 20 - 40 people per image, the original network achieves an accuracy of 0.896$\pm$0.090 while PFCASA, SimAM and PFCA  achieve accuracies of 0.910$\pm$0.065, 0.903$\pm$0.101, and 0.908$\pm$0.085 respectively. The highest accuracy is achieved by SimAM for images with 40-60 people at 0.939$\pm$0.059. Averaging across all images with fewer than 100 people, the original network achieves an accuracy of 0.914$\pm$0.072 while SimAM and PFCA achieve 0.920$\pm$0.075 and 0.920$\pm$0.069 respectively.
In images between 100 and 500 people PFCA shows an accuracy of 0.933$\pm$0.058 compared to the original network's 0.921$\pm$0.062, SimAM's 0.923$\pm$0.066 and PFCASA's 0.928$\pm$0.062.

\section{Discussion}\label{sec:discussion}

Both MAE and Accuracy show that parameter free attention mechanisms improve the model’s performance for low–medium numbers of people. In particular, our novel combination of spatial and channel attention (PFCASA) shows the highest improvement of $1.2\%$ compared to the original model for images with less than 40 people and PFCA shows the highest accuracy for higher crowd densities, up to 500 people, with an improvement of $1\%$. Their respective strengths could be attributed for the former to the joint spatial–channel attention design, which enhances the localization of salient regions while preserving inter-channel feature relationships, an advantage in sparse scenes with isolated individuals and high background variability. For the latter, as crowd density increases, spatial cues become less informative due to severe overlap and occlusion. In such cases, channel attention mechanisms like PFCA prove more effective, as they focus on inter-channel dependencies rather than spatial localization, enabling the model to maintain discriminative capability even when spatial separation between individuals is minimal.
However, both MSE and the standard deviation of accuracy  highlight how much the model’s performance, in particular for low people count and Part A in general, might depend on specifics in the image itself, such as head orientation, camera angle, severe occlusion, etc.

Nonetheless, PFCASA shows promise in improving model performance, without adding to it's complexity. This makes it particularly suitable for edge applications, such as crowd counting in the field of public transportation, where highly sensitive data has to be processed locally.


\begin{figure}
    \centering
    \includegraphics[width=0.90\linewidth]{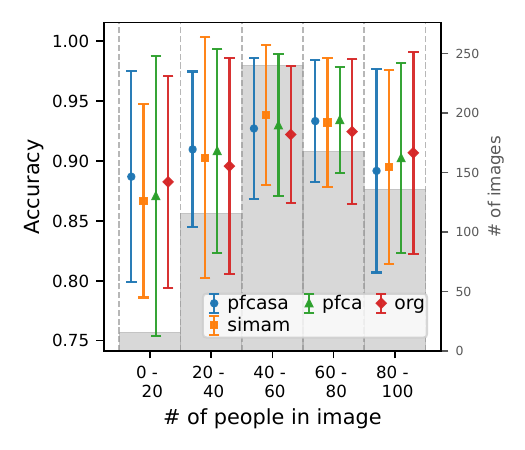}
    \caption{Model performance in terms of accuracy for crowd densities relevant for public transport (0 - 100 people per image) are visualized here. Only the best performing models were selected for visualization.}
    \label{fig:partb_limited}
\end{figure}

\section{Conclusion and future work}\label{sec:conclusion}

 This study investigated the integration of state-of-the-art parameter-free attention mechanisms and a novel attention mechanism into the CSRNet architecture for crowd counting, aiming to enhance accuracy while preserving efficiency for real-time deployment on edge devices. Our experimental results on the ShanghaiTech dataset showed that parameter-free attention modules consistently improved MAE performance in sparse scenes (Part B), supporting their potential applicability in smart public transport modalities. The experiments focused on Part B of the dataset, which is more representative of public transport environments due to its moderate density and varied crowd distribution, indicating that the novel attention mechanism introduced in this paper, PFCASA, achieves superior results in images with fewer than 40 individuals, while PFCA performs better as crowd density increases. Consequently, the parameter-free PFCASA demonstrates the potential to substitute parameterized attention modules considered and be integrated with occupancy detection models on resource-constrained edge devices in intelligent transport systems. In future work, we plan to evaluate the proposed model using a real public transport dataset collected in Linz, Austria, to further validate its effectiveness in practical deployment scenarios.

\section*{Acknowledgment}
This work has been conducted as a part of the OptiPEx project (No.101146513) funded by the European Union. Views and opinions expressed are however those of the author(s) only and do not necessarily reflect those of the European Union or CINEA. Neither the European Union nor the granting authority can be held responsible for them.

\bibliographystyle{IEEEtran}
\bibliography{root} 
	
\end{document}